\documentclass[]{spie}  

\usepackage{amsmath,amsfonts,amssymb}
\usepackage{graphicx}
\usepackage[colorlinks=true, allcolors=blue]{hyperref}
\usepackage{multirow}
\usepackage{color,soul}

\title{Synthesis of Late Gadolinium Enhancement Images via Implicit Neural Representations for Cardiac Scar Segmentation}

\author[a,b,c]{Soufiane Ben Haddou}
\author[a,b,c]{Laura Alvarez-Florez}
\author[d]{Erik J. Bekkers}
\author[e,f]{Fleur V. Y. Tjong}
\author[e,f]{Ahmad S. Amin}
\author[e]{Connie R. Bezzina}
\author[a,b,c,g,h]{Ivana Išgum}

\affil[a]{Department of Biomedical Engineering and Physics, Amsterdam UMC, The Netherlands}
\affil[b]{QurAI group, Informatics Institute, University of Amsterdam, The Netherlands}
\affil[c]{Amsterdam Cardiovascular Sciences, Amsterdam UMC, The Netherlands}
\affil[d]{Amsterdam Machine Learning Lab, University of Amsterdam, The Netherlands}
\affil[e]{Department of Experimental Cardiology, Amsterdam Cardiovascular Sciences, Heart Failure \& Arrhythmias, Amsterdam UMC, The Netherlands}
\affil[f]{Department of Cardiology, Amsterdam UMC, The Netherlands}
\affil[g]{Department of Radiology and Nuclear Medicine, Amsterdam UMC, The Netherlands}
\affil[h]{Department of Radiology, Mayo Clinic, Rochester, United States of America}

\authorinfo{Corresponding author: Soufiane Ben Haddou (s.benhaddou@amsterdamumc.nl)}

\begin{document} 
\maketitle

\begin{abstract}
Late gadolinium enhancement (LGE) imaging is the clinical standard for myocardial scar assessment, but limited annotated datasets hinder the development of automated segmentation methods. We propose a novel framework that synthesises both LGE images and their corresponding segmentation masks using implicit neural representations (INRs) combined with denoising diffusion models. Our approach first trains INRs to capture continuous spatial representations of LGE data and associated myocardium and fibrosis masks. These INRs are then compressed into compact latent embeddings, preserving essential anatomical information. A diffusion model operates on this latent space to generate new representations, which are decoded into synthetic LGE images with anatomically consistent segmentation masks. Experiments on 133 cardiac MRI scans suggest that augmenting training data with 200 synthetic volumes contributes to improved fibrosis segmentation performance, with the Dice score showing an increase from 0.509 to 0.524. Our approach provides an annotation-free method to help mitigate data scarcity.The code for this research is publicly available.\footnote{\url{https://github.com/SoufianeBH/Paired-Image-Segmentation-Synthesis}}

\end{abstract}

\keywords{Late gadolinium enhancement, implicit neural representations, diffusion models, cardiac segmentation, data augmentation}

\section{INTRODUCTION}
\label{sec:intro}  

Cardiac magnetic resonance (CMR) imaging with late gadolinium enhancement (LGE) has emerged as the standard for detecting and quantifying myocardial scars. The gadolinium contrast agent accumulates in areas of myocardial fibrosis, appearing as bright regions that enable visualisation of scar tissue in LGE, providing critical information for patient treatment and  prognosis\cite{polacin2021segmental,jada2024quantification}. However, LGE images do not visualize cardiac anatomy well and inherently suffer from low spatial resolution, motion artifacts, and variable contrast enhancement that challenge expert manual and automatic segmentation algorithms\cite{turkbey2012differentiation}.

The scarcity of high-quality annotated LGE datasets further challenges automatic methods for LGE analysis. 
Manual annotation of cardiac scars requires specialized expertise. 
Accurate delineation typically demands slice wise inspection by medical experts, making the process labor-intensive and subject to inter- and intra-observer variability.
Additionally, scar tissue often exhibits diffuse, heterogeneous appearance with poorly defined boundaries, further complicating accurate delineation.
This data limitation severely constrains the development of robust deep learning models for automated scar quantification.

To address this, traditional LGE image augmentation techniques have been employed \cite{Shorten2019SurveyDataAugmentation,Litjens2017SurveyMedicalImaging}, but they often fail to capture the complex spatial relationships and subtle intensity variations characteristic of myocardial scars visible in LGE images \cite{Chen2022DeepLearningMedicalImage}. Hence, efforts have been made to create synthetic LGE images directly from non-contrast cine CMR sequences by domain translation techniques that learn to match higher-level structural features between the two modalities \cite{campello2020combining}. Although these methods have demonstrated improvement over traditional data augmentation techniques for training automatic LGE segmentation, they still require manual annotations of the generated images and may be limited by the number of available cine CMR scans needed for the synthesis. Hence, approaches to generate synthetic LGE images that can effectively augment existing datasets and improve the performance of current segmentation models are needed.


To address data scarcity in cardiac scar segmentation, we propose a unified framework that synthesises both LGE images and their corresponding segmentation masks using implicit neural representations (INRs) combined with denoising diffusion models. Our approach simultaneously generates anatomically consistent image-mask pairs, eliminating the need for manual annotation of synthetic data. Joint segmentation of myocardium and its scar enables accurate scar localisation within the myocardium and its precise quantification, which are essential for clinical assessment. Beyond improving segmentation performance, the quality and anatomical consistency of synthesized images are critical for their effective use in downstream analysis tasks. In particular, synthetic LGE data must preserve global myocardial geometry, realistic scar localization, and volumetric coherence across slices, as violations of these properties can introduce bias during training. While recent generative approaches have shown promise in medical image synthesis, ensuring alignment between synthesized images and their corresponding segmentation masks remains challenging\cite{turkbey2012differentiation}. Motivated by this, we focus not only on achieving downstream segmentation accuracy, but also on analyzing the anatomical plausibility and structural fidelity of the generated image–mask pairs.

\section{Dataset}
\label{sec:deep_risk_dataset}
This study includes a retrospectively collected set of short-axis LGE CMR images of 133 patients (20-81 years of age, 81\% men) randomly selected from a set of 1261 patients who underwent implantable cardioverter-defibrillator placement at the Amsterdam UMC between 2007 and 2017.  All LGE scans were performed on 1.5T Siemens systems using standard ECG-gated, breath-hold, inversion-recovery segmented gradient-echo sequences, with repetition times (TR) between 430--1138\,ms, echo times (TE) of 1.0--4.4\,ms, and in-plane pixel sizes ranging from 1.33--2.08\,mm. The through-plane spacing ranges from 8.02--10.07\,mm. The images were typically acquired 10--20 minutes after contrast injection. Use of this dataset was approved by the institution's medical ethics committee.

To define the reference standard, two trained annotators manually annotated myocardium and fibrosis pixel-wise, with final verification by an expert cardiologist. 

\section{METHODS \& EXPERIMENTS}
\begin{figure}[t]
    \centering
    \includegraphics[width=1\textwidth]{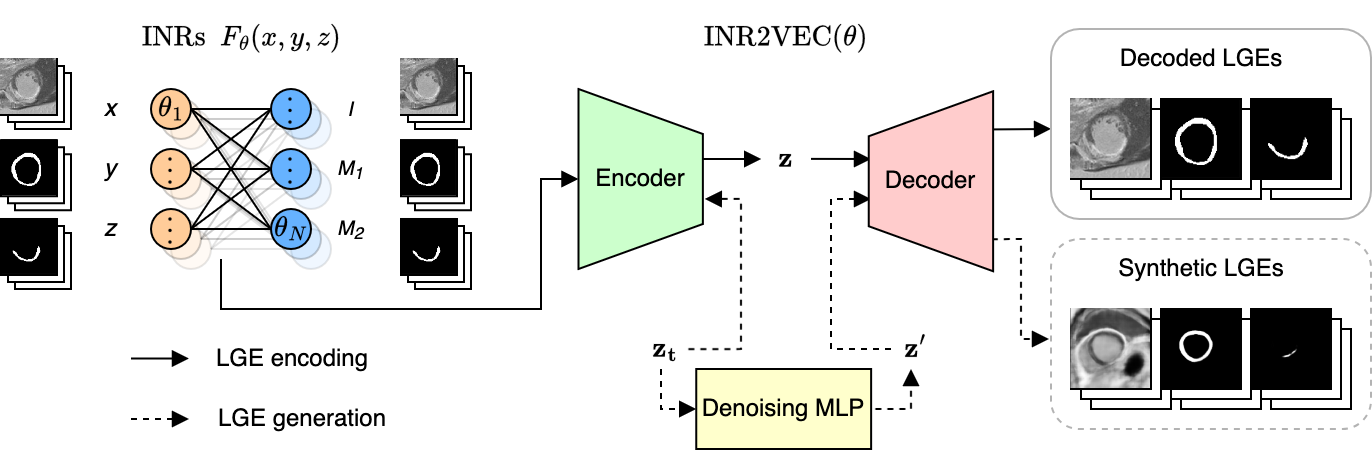}
    \caption{Overview of our proposed framework. An INR model captures both LGE image intensities and its segmentation masks, which are then compressed into a latent space. A diffusion model operates on this latent space to generate synthetic LGE data with anatomically consistent segmentation masks.}
    \label{fig:overview_pipeline}
\end{figure}

To tackle the scarcity of annotated LGE data for training automatic myocardial scar segmentation, we propose an integrated framework that leverages INRs to model LGE images and their corresponding segmentation masks as illustrated in Fig \ref{fig:overview_pipeline}. The learned INR parameters are subsequently compressed into a compact latent space using INR2VEC\cite{de2023deep}, facilitating efficient data manipulation. Finally, a diffusion-based generative model operates within this latent space to synthesize new image–masks sets. These synthetic data are then incorporated into the training set for a deep learning segmentation.


\subsection{Joint Image-Mask Implicit Neural Representation} We model LGE images and their segmentation masks jointly within a single implicit neural representation framework to ensure anatomical consistency. As shown in Fig \ref{fig:overview_pipeline}, INRs capture both image and mask information that is subsequently compressed and used for synthesis by diffusion models to augment training data. We model LGE images and their segmentations as continuous functions using Sinusoidal Representation Networks (SIREN)\cite{sitzmann2020siren} with periodic activations that excel at capturing fine details. Our architecture employs 4 layers with 256 hidden units each. The INR maps 3D spatial coordinates $\mathbf{x}=[x,y,z]^T$ to multiple outputs: $F_{\boldsymbol{\theta}}(\mathbf{x}) \rightarrow (\hat{I}(\mathbf{x}), \hat{M}_1(\mathbf{x}), \hat{M}_2(\mathbf{x}))$, where $I(\mathbf{x})$ represents grayscale image intensity, $M_1(\mathbf{x})$ and $M_2(\mathbf{x})$ represent binary masks for myocardium and fibrosis, respectively. Our multi-task architecture leverages anatomical correlations through a shared backbone network $\phi_{\text{shared}}$ that extracts common features, followed by task-specific heads $\psi_{\text{task}}$ for each output: $
F_{\boldsymbol{\theta}}(\mathbf{x}) = \psi_{\text{task}}(\phi_{\text{shared}}(\mathbf{x}))$. This design enables implicit weight-sharing for anatomical structures while allowing specialized learning for intensity and segmentation predictions. We optimize the network using a composite loss function\cite{molaei2023implicit}:

\begin{equation}
\mathcal{L}(\boldsymbol{\theta}) = \frac{1}{|\Omega|}\sum_{\mathbf{x}\in\Omega}\left[\|I(\mathbf{x})-\hat{I}(\mathbf{x})\|_2 + \text{BCE}(M_1,\hat{M}_1) + \text{BCE}(M_2,\hat{M}_2)\right]
\end{equation} where $\|\cdot\|_2$ denotes L2 norm for image reconstruction and BCE (binary cross-entropy) for mask predictions. \\ \\

\subsection{Latent Space Compression and Generation} Direct manipulation of INR parameters is computationally infeasible due to high dimensionality. We employ INR2VEC\cite{de2023deep} to compress the flattened parameter vector $\boldsymbol{\theta}=[\theta_1,...,\theta_N]^T$ into a compact latent representation: $\mathbf{z} = \text{INR2VEC}(\boldsymbol{\theta}) \in \mathbb{R}^{512}$

The encoder consists of 4 linear layers with batch normalisation, ReLU activations, and max pooling, reducing dimensionality from $\sim 260\text{K}$ parameters to a 512-dimensional latent vector while preserving essential anatomical information. \\ \\
\subsection{Diffusion-based Synthesis} To generate diverse and high-quality synthetic cardiac MRI data, we require a generative model that can capture the complex distribution of anatomical variations in our latent space. Diffusion models have emerged as state-of-the-art generative models, surpassing GANs in both sample quality and training stability while avoiding mode collapse. Their ability to model complex, multi-modal distributions makes them particularly suitable for medical imaging, where anatomical structures exhibit subtle variations that must be preserved. We apply the Elucidated Diffusion Model (EDM) \cite{karras2022elucidated} to generate new latent representations. The forward diffusion process progressively adds Gaussian noise over $T=1000$ timesteps:
\begin{equation}
\mathbf{z}_t = \sqrt{1-\beta_t}\mathbf{z}_{t-1} + \sqrt{\beta_t}\boldsymbol{\epsilon}, \quad \boldsymbol{\epsilon}\sim\mathcal{N}(0,I)
\end{equation}
The reverse process learns to denoise using a neural network $\boldsymbol{\epsilon}_{\boldsymbol{\theta}}$:
\begin{equation}
\hat{\mathbf{z}}_{t-1} = \frac{1}{\sqrt{1-\beta_t}}\left(\mathbf{z}_t - \sqrt{\beta_t}\boldsymbol{\epsilon}_{\boldsymbol{\theta}}(\mathbf{z}_t,t)\right)
\end{equation}
Synthetic data generation involves: (1) sampling $\mathbf{z}_T \sim \mathcal{N}(0,I)$; (2) iterative denoising to obtain $\mathbf{z}_0$; (3) decoding via INR2VEC decoder, and (4) querying the decoded INR to generate images and masks: $(I',M_1',M_2')=\text{Decoder}(\mathbf{z}')(\mathbf{x})$.

\subsection{Image segmentation}
To evaluate whether our synthesized LGE images and their segmentations are beneficial in training, we perform automatic myocardium and scar segmentation. In our segmentation experiments we use nnU-net, current state-of-the-art segmentation network. We use its default configuration, with the only variable being the number of synthetic volumes added to the training set.

\subsection{Experimental Setup}
We divided our dataset into 105 volumes for INR training and 28 for held-out testing. The INR2VEC model was trained on all 105 fitted INRs. The diffusion model was trained on latent vectors from the 105 training cases. For segmentation, we trained three nnU-Net\cite{isensee2018nnu} models with increasing numbers of synthetic samples for augmentation: (1) baseline with 105 real volumes only, (2) baseline augmented with 100 synthetic volumes, and (3) baseline augmented with 200 synthetic volumes.

\section{RESULTS}
We first qualitatively and quantitatively evaluate synthesized LGE images and their segmentation masks. Then, we investigate the impact of augmenting existing datasets with synthetic samples on myocardium and fibrosis segmentation performance.

\begin{figure}[t]
    \centering
    \begin{minipage}[b]{.7\textwidth}
        \centering
        \includegraphics[width=\linewidth]{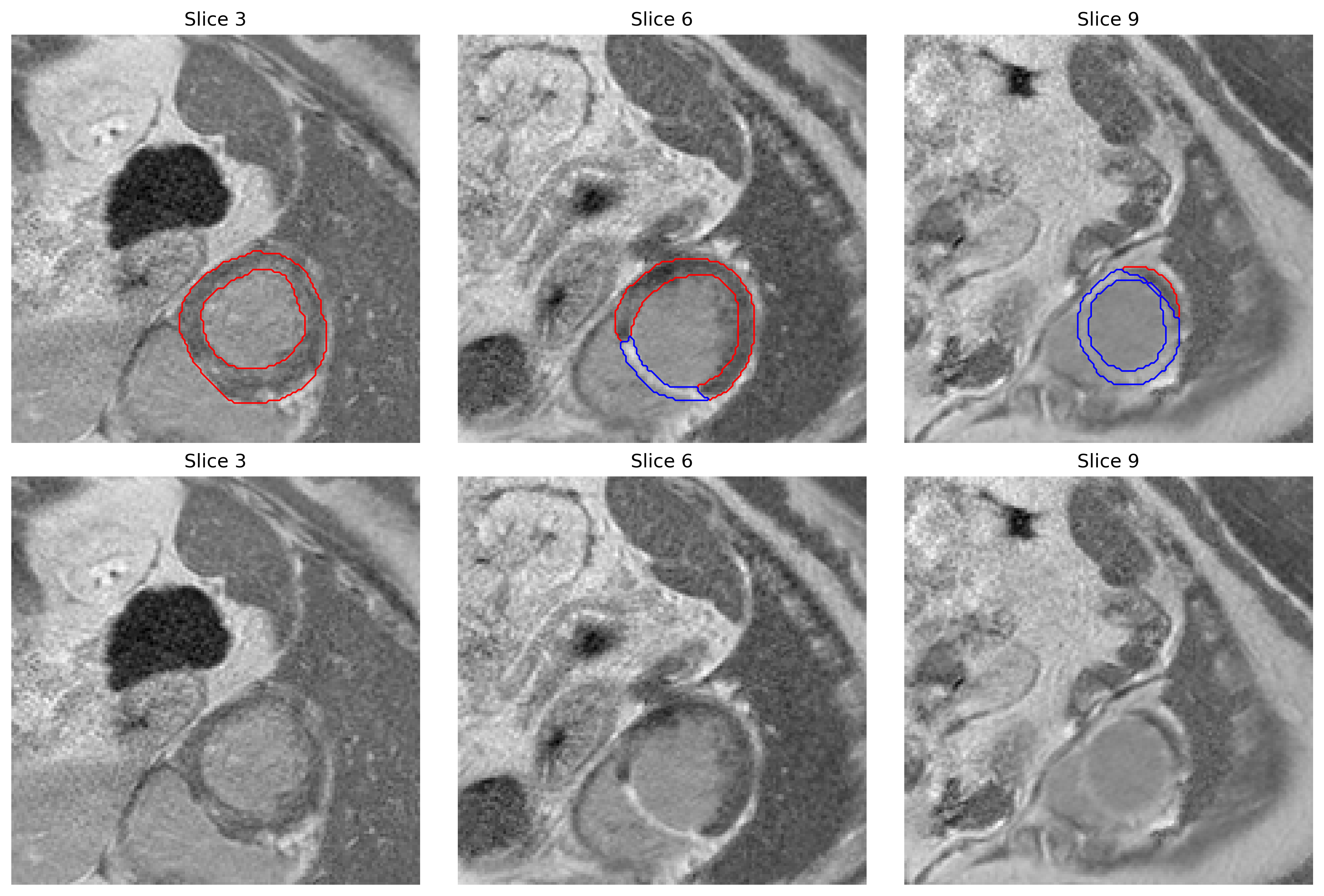}
        (a) Real LGE slices
    \end{minipage}
    \begin{minipage}[b]{.7\textwidth}
        \centering
        \includegraphics[width=\linewidth]{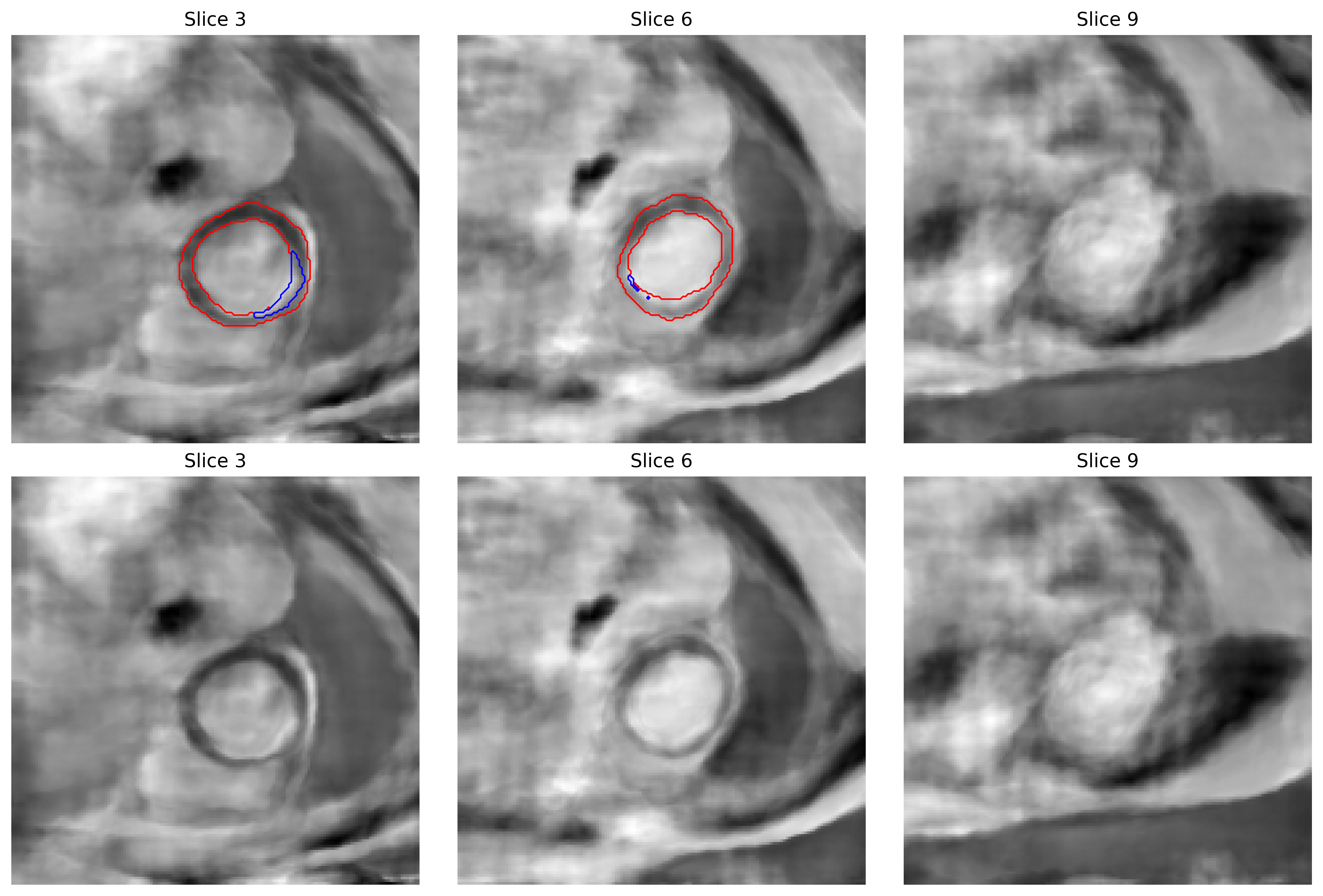}
        (b) Synthetic LGE slices
    \end{minipage}
    \caption{Comparison of real (top) and an example synthetic (bottom) LGE short-axis slices at basal (slice 3), mid-ventricular (slice 6), and apical (slice 9) levels. Top: Slices with ground truth (a) or generated (b) segmentation contours (red = myocardium, blue = fibrosis). Bottom: Same slices without overlay showing image quality and anatomical consistency.}
    \label{fig:synthesis}
\end{figure}

\begin{figure}[t]
    \centering
    \includegraphics[width=1\textwidth]{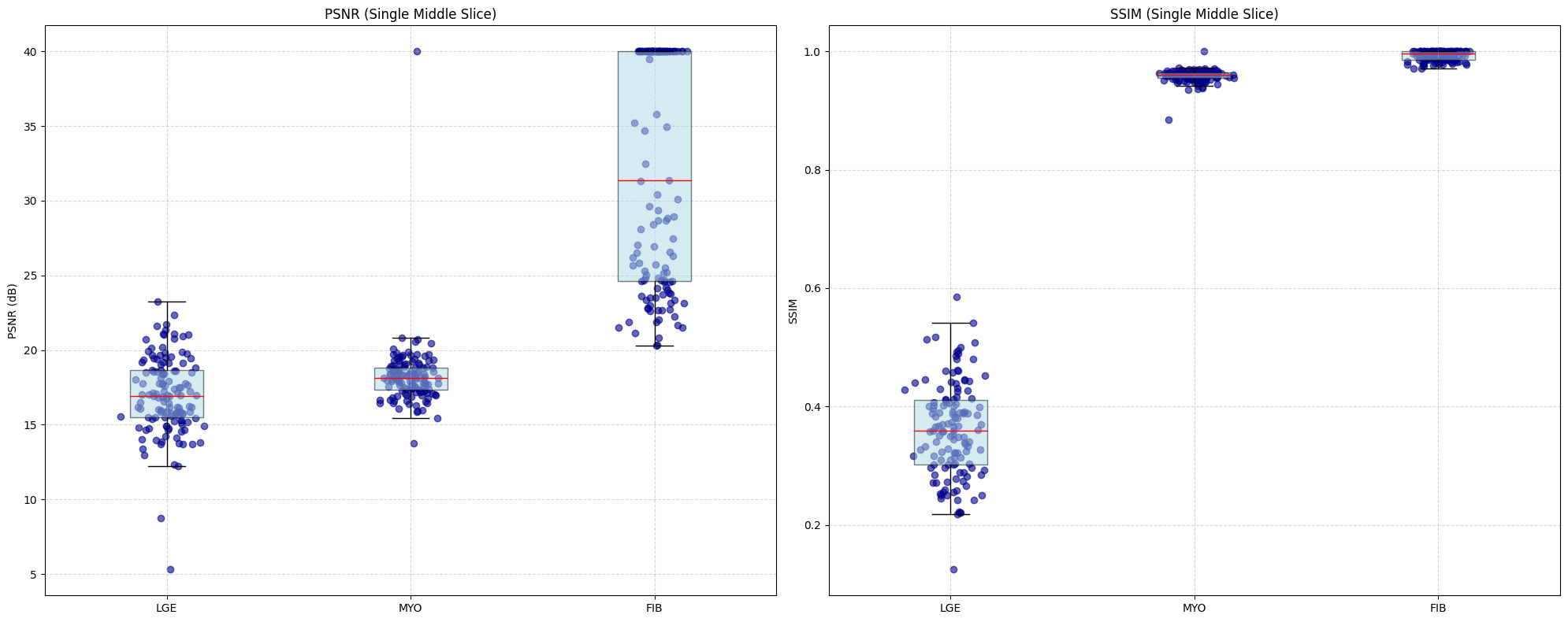}
    \caption{PSNR and SSIM for synthetic vs.\ real samples, evaluated on the middle LGE slice from each volume.}
    \label{fig:psnr_ssim}
\end{figure}

\begin{figure}[t]
\centering

\begin{minipage}{0.49\textwidth}
  \centering
  \includegraphics[width=0.48\textwidth,height=3.4cm,keepaspectratio]{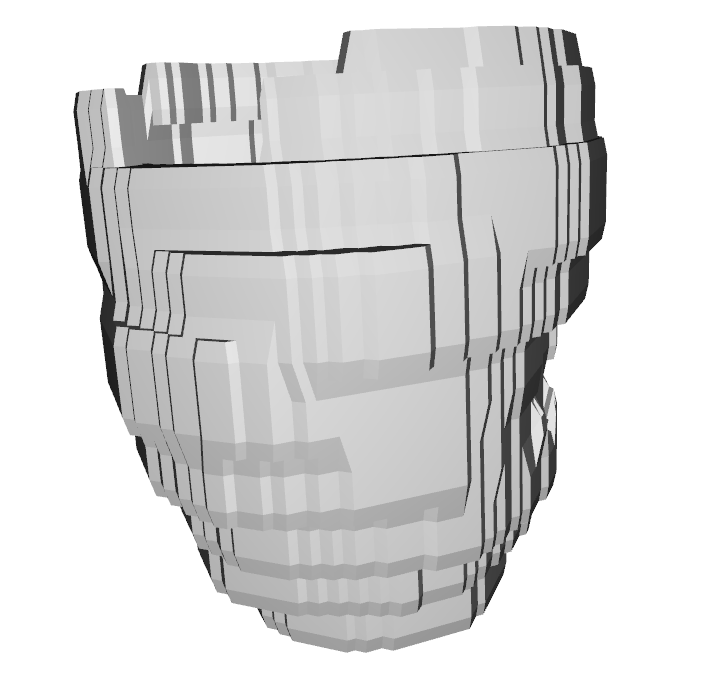}
  \includegraphics[width=0.48\textwidth,height=3.4cm,keepaspectratio]{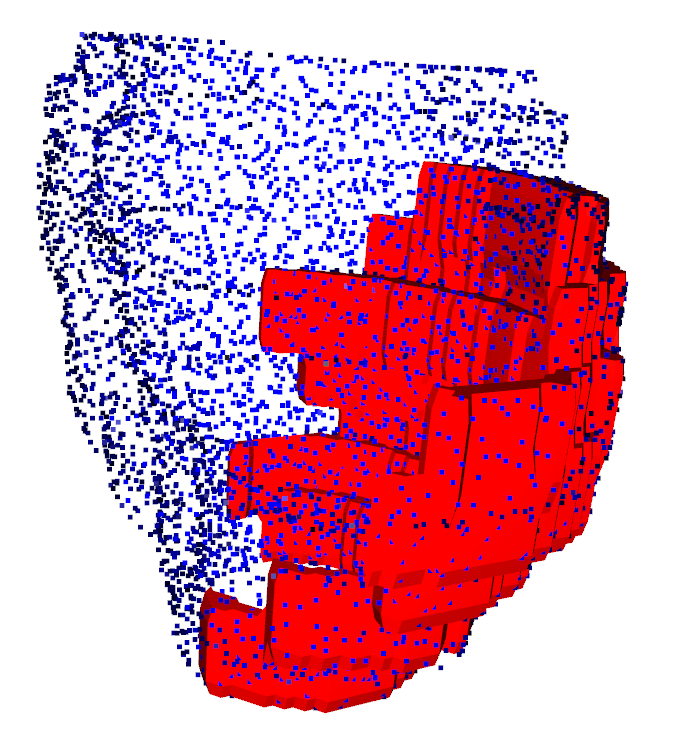}\\
  (a) Real LGE geometry
\end{minipage}\hfill
\begin{minipage}{0.49\textwidth}
  \centering
  \includegraphics[width=0.48\textwidth,height=3.4cm,keepaspectratio]{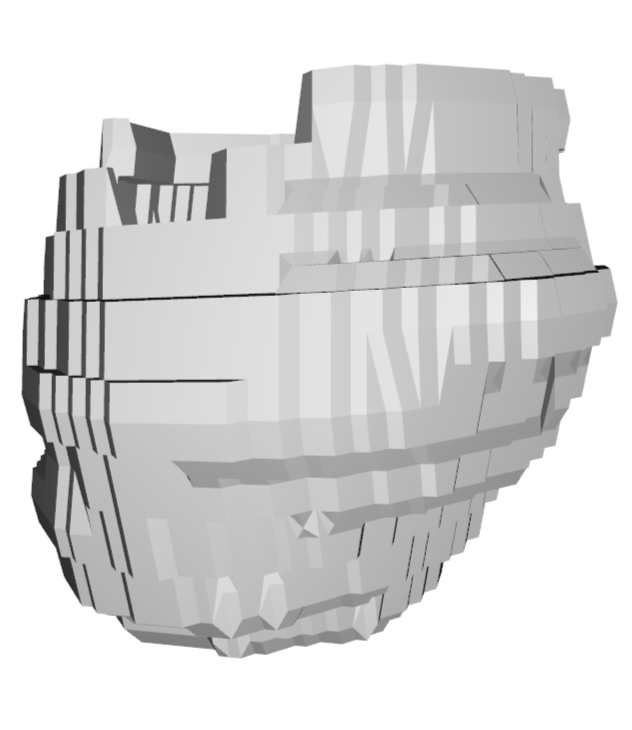}
  \includegraphics[width=0.48\textwidth,height=3.4cm,keepaspectratio]{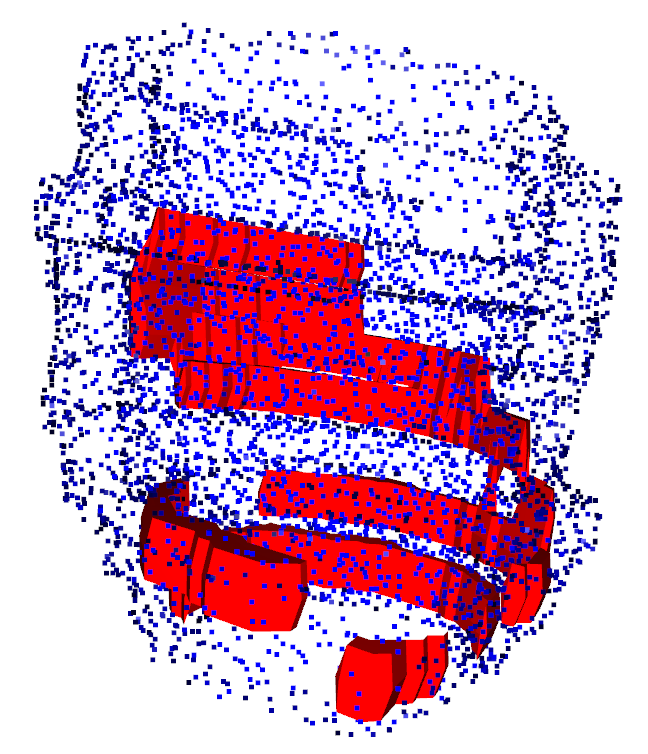}\\
  (b) Synthetic LGE geometry
\end{minipage}

\caption{Comparison of myocardium and fibrosis mask geometries. (a) Real geometry displayed as a grey mesh, with myocardium in blue and fibrosis in red. (b) Synthetic geometry shown using the same representation.}
\label{fig:3d_meshes}
\end{figure}

\subsection{Evaluation of Synthesized LGE Images}
Figure~\ref{fig:synthesis} shows representative examples of real and synthetic LGE short-axis slices at basal, mid-ventricular, and apical levels, with corresponding segmentation contours. Visual inspection shows that the proposed synthesis framework preserves critical anatomical structures relevant for downstream analysis. In particular, the myocardium maintains consistent wall thickness, smooth contours, and coherent geometry across slices, indicating that the latent representation captures global cardiac shape. Fibrosis regions, despite their sparse and heterogeneous appearance in the real data, are generated at anatomically plausible locations within the myocardial wall, i.e., they exhibit realistic spatial relationships relative to surrounding tissue.

While fine-grained texture patterns outside the myocardium are simplified in the synthesized images, this effect is largely confined to regions of limited clinical relevance for scar segmentation. In contrast, areas within the myocardium and fibrosis masks retain higher structural fidelity, which is critical for training segmentation models. This behavior is consistent with the joint image-mask representation learned by the INR, where anatomical structures that are explicitly supervised during training are more accurately preserved than background regions.

To further quantify the similarity between real and synthesized data,
Figure~\ref{fig:psnr_ssim} presents the peak signal-to-noise ratio (PSNR) and structural similarity index (SSIM) computed between synthetic and real LGE samples. As expected, synthetic volumes do not replicate specific real scans on a voxel-level, resulting in moderate PSNR and SSIM values. However, these metrics provide a useful indication of overall structural similarity and confirm that the synthesized images occupy a comparable intensity and texture distribution to the real LGE data. Notably, higher similarity is observed within the myocardial and fibrosis regions compared to surrounding tissue, reflecting the method's focus on the area within these masks.

\subsection{Volumetric and Geometric Consistency}
By visually examining 2D slices (\figurename~\ref{fig:synthesis}) and 3D volumetric representations (\figurename~\ref{fig:3d_meshes}), we observe that the proposed framework produces anatomically plausible cardiac structures across the short-axis stack. The synthesized myocardium maintains consistent wall thickness and coherent geometry across slices, while fibrotic regions are generated at anatomically realistic locations within the myocardial wall.

\subsection{Analysis of the Automatic Segmentations}

We trained a segmentation model on real and synthetic volumes that we then use to segment unseen real image volumes. Table~1 reveals that augmentation with synthetic training data affects fibrosis and myocardium
segmentation differently. Fibrosis segmentation, which is of primary clinical
interest, improves from a baseline Dice coefficient of $0.509 \pm 0.08$ to
$0.524 \pm 0.07$ (+2.9\%) when augmenting the training set with 200 synthetic
volumes. The most pronounced improvement is observed in apical slices (top
25\%: +9.5\%), where fibrosis is typically small, spatially heterogeneous, and
underrepresented in the real training data. 


In contrast, myocardium segmentation shows relatively stable performance across
augmentation levels, with a slight decrease in Dice score when increasing the
number of synthetic samples. Additional synthetic samples may introduce mild
smoothing effects at tissue boundaries, which can marginally affect
boundary-sensitive metrics without substantially altering overall anatomical
correctness.




\begin{table}[htb]
\centering
\caption{Segmentation performance (Dice scores) with varying levels of synthetic data augmentation. Best results in bold.}
\label{tab:results}
\begin{tabular}{lcccccc}
\hline
\multirow{2}{*}{\textbf{Region}} & \multicolumn{3}{c}{\textbf{Fibrosis}} & \multicolumn{3}{c}{\textbf{Myocardium}} \\
\cline{2-7}
& \textbf{N=0} & \textbf{N=100} & \textbf{N=200} & \textbf{N=0} & \textbf{N=100} & \textbf{N=200} \\
\hline
Volume & 0.509 & 0.515 & \textbf{0.524} & 0.815 & \textbf{0.815} & 0.811 \\
Bottom 25\% & 0.195 & \textbf{0.202} & 0.198 & \textbf{0.560} & 0.543 & 0.523 \\
Middle 50\% & 0.549 & 0.549 & \textbf{0.554} & 0.833 & \textbf{0.834} & 0.832 \\
Top 25\% & 0.370 & 0.390 & \textbf{0.405} & \textbf{0.784} & 0.781 & 0.779 \\
\hline
\end{tabular}
\end{table}

\section{DISCUSSION}
We presented a novel framework that jointly synthesises LGE images and their corresponding segmentation masks using INRs combined with diffusion models. This approach addresses the critical challenge of data scarcity in cardiac LGE imaging through several key innovations.

By modelling LGE images and their segmentations within the same INR framework, we ensure anatomical consistency, eliminating the need for manual annotation of synthetic data. To the best of our knowledge, this is the first method that utilizes latent diffusion for synthesis of images with their corresponding segmentation masks. The continuous nature of INRs captures fine-grained spatial relationships often missed by discrete augmentation techniques\cite{mildenhall2021nerf,sitzmann2020implicit}. Although our method segments both myocardium and fibrosis, fibrosis quantification is of primary clinical interest as it directly impacts patient treatment.

Qualitative and quantitative evaluation of the synthesized images indicate that, the synthesized LGE images capture essential anatomical characteristics while introducing realistic variability. Rather than aiming for pixel-perfect reconstruction, the proposed framework generates anatomically coherent image--mask
pairs that are well suited for data augmentation in segmentation tasks, where diversity and structural plausibility are more important than exact visual fidelity.

These qualitative characteristics help explain the observed segmentation trends. For fibrosis, which is sparse and spatially variable, segmentation benefits most from synthetic
augmentation, as the diffusion process introduces additional diversity in scar morphology and location while preserving anatomical plausibility. This effect is particularly evident in apical slices, where fibrosis is often underrepresented in the real training data.

Our results show that fibrosis segmentation improved most in apical slices (+9.5\%), where data scarcity is typically most severe. This improvement is clinically meaningful given that fibrosis segmentation remains one of the most challenging tasks in cardiac MRI due to the small, heterogeneous nature of scar tissue. Furthermore, results show that synthesised LGE images exhibit somewhat limited sharpness at the myocardium blood pool boundaries and simplified texture patterns in regions outside the myocardium. Future work could investigate potential improvements in these areas, such as adversarial training to enhance boundary definition and conditioning the diffusion process on clinical parameters for more targeted fibrosis segmentation with the complexity of reconstruction.

In contrast, myocardium segmentation shows limited improvement, likely due to the stable and well-represented nature of myocardial anatomy in the real dataset. Overall, these observations indicate that synthetic data augmentation is most effective for underrepresented and heterogeneous structures such as myocardial fibrosis, with joint image-mask generation ensuring anatomically consistent and meaningful training samples. Moreover, given that myocardium segmentation does not follow the same improvement in segmentation as fibrosis, future work could address why synthetic data differently impacts fibrosis versus myocardium segmentation. This could inform synthesis strategies for other challenging medical imaging tasks where data scarcity is a fundamental constraint.

The baseline segmentation performance observed in this study is consistent with the known difficulty of myocardial scar segmentation in LGE imaging. Reported fibrosis Dice scores in the literature vary widely, typically ranging from 0.47 to 0.86 depending on dataset characteristics, imaging protocols, and patient populations \cite{zhang2019deep, wu2021recent}. For example, automated scar
quantification methods based on 3D convolutional networks report Dice scores of 0.54--0.57 in hypertrophic cardiomyopathy cohorts \cite{fahmy2018automated}, while other approaches achieve values between 0.55 and 0.82 using different architectural and methodological choices \cite{yang2020simultaneous}. Within this context, the observed improvements obtained through synthetic data augmentation are meaningful, particularly given the absence of additional manual annotations and the focus on anatomically consistent image--mask
generation.

In this work, we show that incorporating synthetic images and segmentations improves the performance of automatic segmentation of fibrosis in LGE images. Using nnU-Net as a strong reference, future work could evaluate task-specialized segmentation networks, focusing on whether they can better leverage synthesis-based augmentation, and moreover systematically test low training data settings to quantify how much real annotation effort can be reduced.

\section{CONCLUSION}
We introduced an INR diffusion framework that jointly generates anatomically consistent LGE images and segmentations masks, enabling an increase in the size of the training data set without additional manual labelling. Experiments demonstrate that synthetic augmentation improves automatic fibrosis segmentation, with the largest segmentation gains in more difficult apical regions where data scarcity and anatomical variability are more pronounced. The findings suggest that coupled image-mask synthesis is a promising strategy for increasing training data diversity in small datasets. This lays the basis for further improvements towards synthesis fidelity and reduction of manual labelling effort.

\acknowledgments 
The project was funded by the call HORIZON-EIC-2022-PATHFINDERCHALLENGES-01 "CARDIOGENOMICS" from HORIZON European Innovation Council Grants/ European Commission (DCM-NEXT project; project number 101115416).

\bibliography{MICCAIbibliography} 
\bibliographystyle{spiebib} 

\end{document}